\documentclass[10pt,twocolumn,letterpaper]{article}

\usepackage{iccv}
\usepackage{times}
\usepackage{epsfig}
\usepackage{multirow}
\usepackage{graphicx}
\usepackage{amsmath}
\usepackage{amssymb}

\usepackage{bbm}
\usepackage{multirow}
\usepackage{balance}

\usepackage[pagebackref=true,breaklinks=true,letterpaper=true,colorlinks,bookmarks=false]{hyperref}

\iccvfinalcopy 

\def\iccvPaperID{5418} 

\ificcvfinal\fi

\begin{document}

\title{ECACL: A Holistic Framework for Semi-Supervised Domain Adaptation}
\author{Kai Li$^{1,2}$\thanks{Work done when Kai Li was with Northeastern University.}\hspace{3pt}, Chang Liu$^{2}$, Handong Zhao$^{3}$, Yunlun Zhang$^{2}$, Yun Fu$^{2}$ \\
$^1$NEC Laboratories, America, 
$^2$Northeastern University,
$^3$Adobe Research \\
{\tt\small kaili@nec-labs.com, liu.chang6@northeastern.edu, hdzhao@adobe.com} \\ {\tt\small yulun100@gmail.com, yunfu@ece.neu.edu}
}

\maketitle
\ificcvfinal\thispagestyle{empty}\fi

\begin{abstract}
This paper studies Semi-Supervised Domain Adaptation (SSDA), a practical yet under-investigated research topic that aims to learn a model of good performance using unlabeled samples and a few labeled samples in the target domain, with the help of labeled samples from a source domain. Several SSDA methods have been proposed recently, which however fail to fully exploit the value of the few labeled target samples. In this paper, we propose Enhanced Categorical Alignment and Consistency Learning (ECACL), a holistic SSDA framework that incorporates multiple mutually complementary domain alignment techniques. ECACL includes two categorical domain alignment techniques that achieve class-level alignment, a strong data augmentation based technique that enhances the model's generalizability and a consistency learning based technique that forces the model to be robust with image perturbations. These techniques are applied on one or multiple of the three inputs (labeled source, unlabeled target, and labeled target) and align the domains from different perspectives. ECACL unifies them together and achieves fairly comprehensive domain alignments that are much better than the existing methods: For example, ECACL raises the state-of-the-art accuracy from 68.4 to 81.1 on VisDA2017 and from 45.5 to 53.4 on DomainNet for the 1-shot setting. 
Our code is available at \url{https://github.com/kailigo/pacl}.
\end{abstract}

\section{Introduction}
Domain adaptation investigates techniques of avoiding severe performance drop when deploying a model on a new domain (target) that has domain gap with the one (source) which the model is trained on. Most existing research focuses on Unsupervised Domain Adaptation (UDA) where a model is trained jointly with unlabeled target data and labeled source data. Many effective UDA approaches have been proposed, from early works that project data from both domains to a shared feature space \cite{gong2012cvpr,pan2011tnn}, to recent ones that are based on adversarial learning \cite{DBLP:conf/icml/ChenWLW19,long2018conditional}.

This paper makes a slight diversion from the mainstream UDA research direction and investigates how much it can help if we are further provided with a few (e.g., one sample per class) labeled target samples. We call these scarce labeled target samples as ``landmarks''. This is a practical (with minimal labeling effort) yet under-investigated problem and is referred as semi-supervised domain adaptation (SSDA). Preliminary works  before the deep learning era use  landmarks to more precisely measure the data distribution mismatch between source and target domains, either based on Maximum Mean Discrepancy (MMD) or domain invariant subspace learning \cite{ao2017fast,donahue2013semi,yao2015semi}. Recent ones revisit this problem and establish new evaluation benchmarks in the deep learning context \cite{saito2019semi,kim2020attract}. However, these prior works have not fully realized the value of the precious landmarks: They are mainly used to optimize the cross-entropy loss along with the labeled source samples that are of a much greater amount;
the contribution of the landmarks is significantly diluted and thus the learned model shall still be biased towards the source domain.

In this paper, we propose Enhanced Categorical Alignment and Consistency Learning (ECACL), a SSDA framework which unifies multiple techniques that align domains from different complementary perspectives.  
As we have access to a few labeled samples from the target domain, i.e., landmarks, we can align the domains in a supervised way by explicitly aligning samples of the same category from the two domains. We propose two techniques to achieve this objective.
The first one is based on the prototypical loss \cite{snell2017prototypical,li2019rethinking}. We calculate a target prototype for each class by averaging feature embeddings of the landmarks from that class. Then, source samples are aligned with the target prototype from the same class. 
The second one is based on the triplet loss. We explicitly push source samples close to the landmarks from the same class and apart from the landmarks from the different classes.

Overfitting would likely occur since the number of landmarks is small. An intuitive approach to this is data augmentation. But rather than employing some commonly used techniques, e.g., image flipping, we harvest a recently proposed one, RandAugment \cite{cubuk2019randaugment} which produces highly perturbed images by applying various image transformations on an image.
We apply RandAugment on both labeled source samples and landmarks, which makes the categorical alignment non-trivial to achieve and thus enhances model generalibility.

Consistency learning has recently been proved a very successful solution
to various label scarce problems \cite{sohn2020fixmatch,xie2019unsupervised,chen2020simple}.
Inspired by this, we introduce consistency learning to cope the SSDA problem. Specifically, we apply simultaneously a light data augmentation (e.g., image flipping) and a strong augmentation (e.g., RandAugment) on each unlabeled target image, and obtain two versions for each image. We enforce the consistency constraint by producing a pseudo label from the lightly augmented version, and use the pseudo label as the ground truth label for the strongly augmented version for supervised learning. 
By requiring different perturbed versions of the same image being predicted with the same label, we encourage the model to be robust to changes in the image space and thus be more capable of handling the domain gap. Besides, since unlabeled target samples share the same label space as the labeled source samples, this constraint facilitates label propagation from the labeled source domain to the unlabeled target domain. 

Integrating the above techniques that align domains from different perspectives using different combinations of the inputs, we reach the holistic SSDA framework, ECACL. We show in the experiments that ECACL
significantly advances the state-of-the-art performance on the common evaluation benchmarks.
For example, 
ECACL lifts the state-of-the-art mean accuracy from 68.4 to 81.1 on \textit{VisDA2017} and from 45.5 to 53.4 on \textit{DomainNet} for the 1-shot setting

In summary, the contributions of this paper are as follows: 
(1) We propose ECACL, a holistic SSDA framework that incorporates multiple complementary domain alignment techniques. Although each of the incorporated technique is not fundamentally new, we are the first to introduce them to address the SSDA problem and assemble them in a holistic framework.
(2) We conduct a comprehensive ablation study and analysis of ECACL, which offer insights on drawing connections among seemingly distinct tasks and identifying contributing techniques.
(3) We significantly advance the state-of-the-art performance for SSDA.

\begin{figure*}
\centering
\includegraphics[width=0.95\linewidth]{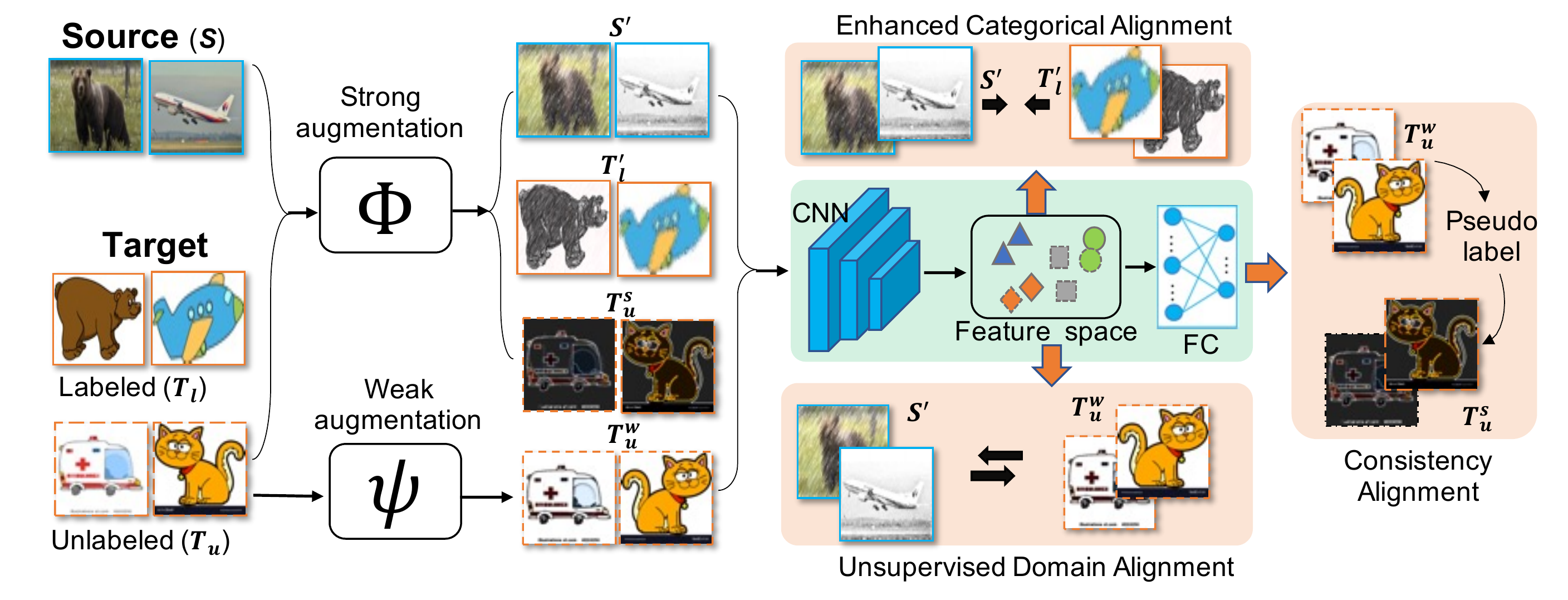}   
\vspace{-5pt}
\caption{Illustration of the proposed ECACL framework. ECACL includes three modules: (1) The unsupervised domain alignment module that performs domain alignment using labeled source and unlabeled target samples. 
(2) The enhanced categorical alignment module that conducts class-level alignment by explicitly pushing close cross-domain images that are from the same class, even when the images are strongly perturbed. (3) The consistency alignment module that generates a pseudo label for each unlabeled target sample from its weakly augmented version and applies the pseudo label on its strongly augmented version with supervised learning.  
}
\label{fremework}  
\vspace{-5pt}
\end{figure*}

\section{Related Work}
\noindent \textbf{Domain Adaptation (DA)}.
According to the type of data available in target domain, DA methods can be divided into three categories: Unsupervised Domain Adaptation (UDA), Few-Shot Domain Adaptation (FSDA) and Semi-Supervised Domain Adaptation (SSDA). UDA assumes that target domain data are purely unlabeled. Early methods in the shallow regime address UDA either by reweighting source instances \cite{huang2007nips,gong2013icml} or projecting samples into a domain invariant feature space \cite{gong2012cvpr,pan2011tnn}. Recent ones are more in the deep regime and approach UDA by moment matching \cite{long2015learning,DBLP:conf/iccv/CarlucciPCRB17,DBLP:conf/icml/LongZ0J17} or adversarial learning \cite{DBLP:conf/icml/ChenWLW19,DBLP:journals/jmlr/GaninUAGLLML16,DBLP:conf/nips/LuoZHL17,long2018conditional,li2020cross}.
FSDA assumes that there is no access to unlabeled samples, but a few labeled ones in target domain. To fully utilize the few labeled target samples, existing methods perform class-wise domain alignment using contrastive loss \cite{motiian2017unified} or triplet loss \cite{xu2019d}.
SSDA is a hybrid of FSDA and UDA where we have access to both a few labeled samples and many unlabeled samples from target domain. 
Early works use the extra labeled target samples to help more precisely measure the data distribution mismatch between source and target domains, either based on Maximum Mean Discrepancy (MMD) or domain invariant subspace learning \cite{ao2017fast,donahue2013semi,yao2015semi,saito2019semi,tejankar2019semi}.  
Saito et al. recently proposed a deep learning based method which alternates between maximizing the classification entropy with respect to the classifier and minimizing it with respect to the 
feature encoder \cite{saito2019semi}. Kim et al. extended this work by alleviating the intra-domain discrepancy problem \cite{kim2020attract}.  We approach SSDA in a new way by proposing a general framework into which existing 
UDA methods can be incorporated as one component for domain alignment along with other novel components.

\noindent \textbf{Semi-Supervised Learning (SSL)}.
Leveraging unlabeled data along with labeled ones in the training process, SSL has boosted performance with a variety of training strategies, including graph-based \cite{kipf2016}, adversarial \cite{miyato2016iclr}, generative~\cite{dai2017nips}, model-ensemble~\cite{laine2016}, self-training~\cite{li2019nips,sohn2020fixmatch}, etc.  The difference of SSL and SSDA lies that labeled samples of SSL are from the same domain as the unlabeled ones. In SSDA, labeled samples instead come from two different domains and the majority are out-of-domain (relative to the target).
So, compared with SSL, SSDA needs first address the domain shift problem in order to leverage the plenty yet out-of-domain labeled samples. We achieve this by employing off-the-shelf UDA techniques and proposing the categorical domain alignment techniques. 


\noindent \textbf{Few-Shot Learning (FSL)}. FSL aims to acquire knowledge of \textit{novel classes} with only a few labeled samples \cite{vinyals2016matching,snell2017prototypical,li2020adversarial,li2019novel}. FSL has very distinct goals from SSDA. FSL emphasizes generalizability of a learned model towards novel classes for which there is no sample (neither labeled nor unlabeled) available during training but a few labeled ones in test. SSDA instead focuses on enhancing generalizability of a model towards unlabeled samples of the classes for which during training there are plenty of labeled samples from source domain, a few labeled ones and many unlabeled ones from target domain. Even with different goals, the way that an FSL method \cite{snell2017prototypical} utilizes a few labeled samples to recognize other unlabeled ones inspires us to develop the supervised alignment module which achieves categorical domain alignment.


\section{Algorithm}
Semi-supervised domain adaptation (SSDA) investigates the adaptation from a label-rich source dataset 
$\mathcal{S}=\{(\textbf{s}_i, y^s_i)\}^{N_s}_{i=1}$ 
to a label-scarce target dataset $\mathcal{T}=\mathcal{T}_l\cup\mathcal{T}_u$, where 
$\mathcal{T}_l=\{\textbf{t}_i, y^t_i\}^{N_t}_{i=1}$ is a labeled set
and $\mathcal{T}_u=\{\textbf{u}_i\}^{N_u}_{i=1}$ an unlabeled set.  
$\mathcal{S}$ and  $\mathcal{T}$
are drawn from the same label space $\mathcal{Y}=\{1, 2, \dotsc,  C\}$ but with different data distributions that cause domain shifts. 
Usually, the number of labeled samples in $\mathcal{T}_l$ is very small, e.g., one sample per class in the extreme case. We call these labeled target samples as ``landmarks''.  Our goal is to learn a domain adaptive model using $\mathcal{S}$, $\mathcal{T}_u$ and landmarks $\mathcal{T}_l$. Let the model be $h=f\circ g$ with parameters $\theta$, where $f$ generates features from images and $g$ outputs label predictions based on the extracted features. 

We can see that the difference of SSDA from UDA is the extra access to landmarks. A naive way to extend an UDA method to SSDA is to optimize
\begin{equation}
\mathcal{L}_{uda} = \mathcal{L}_{ce} + \alpha\mathcal{L}_{ua},
\label{loss_uda}
\end{equation}
where 
\begin{equation}
\mathcal{L}_{ce} = \frac{1}{N_s}\sum_{(\mathbf{s}_i, y^s_i)\sim\mathcal{S}}L(h(\textbf{s}_i), y^s_i) 
+ 
\frac{1}{N_t}\sum_{(\mathbf{t}_i, y_i^t)\sim\mathcal{T}_l}L(h(\textbf{t}_i), y_i^t)
\label{loss_ce}
\end{equation}
is the cross-entropy loss over labeled source and target samples. $\mathcal{L}_{ua}$ is an UDA loss that exploits unlabeled target samples and labeled source samples for domain alignment. It varies in different UDA methods.


Naively merging landmarks into source samples for the cross-entropy loss optimization does not fully release their potentials, as their contribution would be severely diluted. The learned model would thus still be biased towards the source domain. 
We solve this by performing \textit{categorical alignment} where we explicitly
push source samples towards the landmarks from the same class and apart from the landmarks from the different classes (Section \ref{sec:ca}). This encourages the model to produce features that maintain class discrimination despite of domain shifts.
In Section \ref{sec:da}, we further enhance the categorical alignment technique with 
a data augmentation strategy where images are heavily perturbed, making the task harder to fulfill and hence improving model generalizability.
Besides, we carefully design another data augmentation based technique that is applied on unlabeled target samples. This technique constrains the model to make consistent predictions for different versions of the same image undergone different levels of perturbations.
Figure \ref{fremework} shows our framework.

\subsection{Categorical Alignment}
\label{sec:ca}
With labeled source samples $\mathcal{S}=\{(\textbf{s}_i, y^s_i)\}^{N_s}_{i=1}$  and  labeled target samples
$\mathcal{T}_l=\{\textbf{t}_i, y^t_i\}^{N_t}_{i=1}$, we propose two approaches to achieve categorical alignment, one based on the prototypical loss and the other based on the triplet loss.

\noindent\textbf{Prototypical loss based approach}.
This approach learns to minimize the risk of assigning source samples to the landmarks from the same class. Specifically, we calculate a target representation, or a target prototype  for each class by averaging the feature embeddings of the landmarks from that class \cite{snell2017prototypical}:
\begin{equation}
\mathbf{c}_k=\frac{1}{|\mathcal{T}_k|}\underset{(\textbf{t}_i, y_i)\in \mathcal{T}_k}{\sum} f(\textbf{t}_i),
\label{prototype}
\end{equation}
where $\mathcal{T}_k$ is the landmark collection for class $k$. With the target prototypes for all classes $\{\mathbf{c}_k\}^C_{k=1}$, we can compute a distribution over classes for a source sample $(\textbf{s}_i, y^s_i)$ based on a softmax over distances to the target prototypes in the embedding space:
\begin{equation}
p(y_i^s=y|\textbf{s}_i) = \frac{\exp(-\|f(\mathbf{s}_i)-\mathbf{c}_y\|_2)}{\sum_{k=1}^C \exp(-\|f(\mathbf{s}_i)-\textbf{c}_k\|_2)}.
\label{probability}
\end{equation}
Then, we can calculate the prototypical loss over all source samples as
\begin{equation}
L_{pa}=\frac{1}{CN_s}\sum_{(\mathbf{s}_i, y_i^s)\sim\mathcal{S}}\sum_{y\sim\mathcal{Y}}y_i^s\log [-p(y^s_i=y|\mathbf{s}_i)].
\end{equation}

Note that although class prototype based solution has been proposed to address UDA \cite{pan2019transferrable,xu2020cross}, we are the first to introduce it for SSDA. Besides, while the class prototypes in the UDA method are calculated based on pseudo labels which are not always reliable \cite{zou2019confidence}, we compute class prototypes directly from landmarks.

\noindent\textbf{Triplet loss based approach}.
This approach explicitly optimizes the model to produce features such that cross-domain samples of the same class should be of higher similarity than those from different classes \cite{hermans2017defense}. Specifically,
for each landmark $(\mathbf{t}_i, y_i^t)\in\mathcal{T}_l$, we find from $\mathcal{S}$ the least similar source sample  $(\textbf{s}_p, y_p)$ which also belongs to class $y_t$ (i.e., the hard positive). Meanwhile, we find from $\mathcal{S}$ the most similar sample $(\textbf{s}_n, y_n)$ which belongs to a class different from $y_t$ (i.e., the hard negative). 
With the hardest triplet $(\mathbf{t}_i, \textbf{s}_n, \textbf{s}_p)$, we optimize the following triplet loss as:
\begin{equation}
\begin{array}{cl}
L_{ta} = & \frac{1}{N_t}\underset{(\mathbf{t}_i, y_t)\sim\mathcal{T}_l}{\sum} \big[\|f(\textbf{t}_i) - f(\textbf{s}_p)\|_2^{2} - \\ & \|f(\textbf{t}_i)-f(\textbf{s}_n)\|_2^2+m\big]_{+}.
\end{array}
\label{triplet_loss}
\end{equation}
The loss makes sure that for any given landmark, its hard positive sample should be closer to it than its hard negative sample  by at least a margin $m$.

\subsection{Domain Alignment with Data Augmentation}
\label{sec:da}
To further alleviate the scarcity of landmarks, we propose 
two data augmentation based domain alignment techniques, one applied on labeled samples and the other applied on unlabeled samples. 

\subsubsection{Enhanced Categorical Alignment}
It has been shown recently that strong augmentation that creates highly perturbed images brings significant performance gains for supervised learning \cite{cubuk2019autoaugment,cubuk2019randaugment}. 
Inspired by this, we introduce strong data augmentation to address the DA problem. 
For each labeled sample from source and target domains, $(\mathbf{s}_i, y^s_i)\in\mathcal{S}$ or $(\mathbf{t}_i, y^t_i)\in\mathcal{T}_l$, we process it with RandAugment  \cite{cubuk2019randaugment} by applying random augmentation techniques sampled from a transformation set, including color, brightness, contrast adjustments, rotation, polarization, etc. This is then followed by the Cutout \cite{devries2017improved}. 
Now we obtain $\mathcal{S}'$ and $\mathcal{T}'$ that consist of highly perturbed images. With $\mathcal{S}'$ and $\mathcal{T}'$, we can get the enhanced categorical alignment objectives. For the prototypical loss based one, we reach 
\begin{equation}
L'_{pa}=\frac{1}{CN_s}\sum_{(\mathbf{s}'_i, y_i^s)\sim\mathcal{S}'}\sum_{y\sim\mathcal{Y}}y_i^s\log [-p(y^s_i=y|\mathbf{s}'_i)],
\label{proto_loss_aug}
\end{equation}
where $P(y_s=y|\mathbf{s}')$ is calculated with Eq. (\ref{probability}), and the target prototypes are calculated with $\mathcal{T}'$.
Similarly, for the triplet loss based one, we have 
\begin{equation}
\begin{array}{cl}
L'_{ta} = & \frac{1}{N_t}\underset{(\mathbf{t}_i', y_t)\sim\mathcal{T}'_l}{\sum} \big[\|f(\textbf{t}'_i) - f(\textbf{s}'_p)\|_2^{2} - \\ & \|f(\textbf{t}'_i)-f(\textbf{s}'_n)\|_2^2+m\big]_{+},
\end{array}
\label{triplet_loss_aug}
\end{equation}
where $\textbf{s}'_p$ and $\textbf{s}'_n$ are the hard positive sample and hard negative sample mined from $\mathcal{S}'$, respectively.

Strong data augmentation produces a wider range of highly perturbed images, which makes the model harder to memorize the few landmarks and therefore enhances the generalizability of the learned model. On the one hand, the model is forced to be insensitive to more diverse changes or perturbations in the image space, which helps domain alignment as these changes model a wide range of factors causing domain shift. 
On the other hand, the above categorical alignment techniques in essence optimize the model extracting image features that the intra-class ones are of higher similarity than the inter-class ones regardless of domain shifts. It is harder for the model to achieve this optimization objective with highly perturbed images. Thus, the model is encouraged to mine the most discriminative class semantics out of highly perturbed images. 





\subsubsection{Consistency Alignment}
Inspired by the recent success of consistency learning in 
semi-supervised learning \cite{sohn2020fixmatch,xie2019unsupervised,berthelot2019remixmatch,berthelot2019mixmatch}, we introduce consistency learning to address SSDA and propose the CONsistency Alignment (CONA) module.
For each unlabeled target sample $\textbf{u}_i\in \mathcal{T}_u$, we apply weak augmentation $\psi$ and strong augmentation $\Phi$:
\begin{eqnarray}
\textbf{u}^w_i & = & \psi(\mathbf{u}_i), \\
\textbf{u}^s_i & = & \Phi(\mathbf{u}_i).
\end{eqnarray}
The weak augmentation $\psi$ includes the widely-used image flipping and image translation. Same as the practice for labeled samples, we use RandAugment \cite{cubuk2019randaugment} and Cutout \cite{devries2017improved} as our strong data augmentation $\Phi$. We feed $\textbf{u}^s_i$ and $\textbf{u}^w_i$ to the model $h$, and optimize the following objective function:
\begin{equation}
L_{cona} = \underset{{\mathbf{u}_i\sim\mathcal{U}}}{\sum} 
\big[\mathbbm{1}(\max(\textbf{p}_w)\geq\sigma) H(\tilde{\textbf{p}}_w, \textbf{p}_s)\big],
\label{self_loss}
\end{equation}
where $\textbf{p}_s$ and $\textbf{p}_w$ are the classification probabilities of augmented samples $\textbf{u}^s_i$ and $\textbf{u}^w_i$, respectively. $\tilde{\textbf{p}}_w=\arg \max(\textbf{p}_w)$ returns a one-hot vector for the prediction;  $H(., .)$ is the cross-entropy of two possibility distributions; $\mathbbm{1}(.)$ is an indicator function and $\max(\textbf{p}_w)$ returns the highest possibility score.

In essence, the above CONA module computes a pseudo label for an unlabeled sample from its weakly-augmented version and applies the pseudo label on its strongly-augmented version for the cross-entropy loss optimization.  To mitigate the impact of incorrect pseudo labels, only the samples with confident predictions (the highest probability scores are above a threshold) are used for loss computation. This introduces a form of consistency regularization, encouraging the model to be insensitive to image perturbations and hence stronger in classifying unlabeled images.

Pseudo labeling (or self-training) has been investigated before for domain adaptation \cite{pan2019transferrable,zou2019confidence}, but our method is clearly distinct from the previous ones. Existing methods usually perform stage-wise pseudo labeling: Each stage consists of a number of training epochs and the latest model is applied on unlabeled samples in the end of each stage. The confidently predicted samples are selected for model training in the next stage usually in the same way as labeled samples from source domain. Within all training epochs in each stage, the pseudo-labeled samples remain unchanged. Our method instead performs mini-batch-wise pseudo labeling: In each mini-batch, we compute a pseudo label for every sample from its weakly-augmented version and apply the pseudo label on the strongly-augmented one. We discard all of the pseudo labels after each mini-batch, which alleviates the harmful impact of incorrect pseudo labels.

The overall learning objective of our method is a weighted combination of the UDA loss, the enhanced categorical alignment loss, and the consistency alignment loss:
\begin{equation}
L = L_{uda} + \lambda_1 L_{cata} + \lambda_2 L_{cona}, \hspace{5pt} L_{cata}=\{L'_{pa}, L'_{ta}\}
\label{obj}
\end{equation}
where $\lambda_1$ and $\lambda_2$ are the hyper-parameters. $L_{cata}=L'_{pa}$ and $L_{cata}=L'_{ta}$ correspond to the two variants of our ECACL framework which performs categorical alignment based on the prototypical loss and the triplet loss, respectively. We refer these two variants as ECACL-P and ECACL-T, respectively. 

\textbf{Algorithm 1} outlines the main steps of the proposed framework.

\begin{table}[t]
  \small
  \centering
  \begin{tabular}{l}
    \hline
    \noindent \textbf{Algorithm 1.} Proposed ECACL framework  \\\hline 
    \textbf{Input:} Source set $\mathcal{S}=\{\mathcal{X}_s, \mathcal{Y}_s\}$, labeled target set  \\ 
    \hspace{9mm} $\mathcal{T}_l=\{\mathcal{X}_t, \mathcal{Y}_t\}$ and unlabeled target set
    $\mathcal{T}_u$. \\
    \textbf{Output:} Domain adaptive model $h$. \\\hline
    \textbf{while} not done \textbf{do} \\ 
    \hspace{1.5mm} 1. Sample from $\mathcal{S}\cup\mathcal{T}_l$ labeled images \\ 
    \hspace{5mm}   $\mathcal{B}_l=\{\{\textbf{s}_{i, j}\}^{N_s}_{i=1}, \{\textbf{t}_{i, j}\}^{N_t}_{i=1}, y_j\}^{M}_{j=1}$. Sample  \\
    \hspace{5mm}  unlabeled images $\mathcal{B}_u=\{\textbf{u}_{i}\}^{N_u}_{i=1}$ from $\mathcal{U}$. Denote  \\ 
    \hspace{5mm} $\mathcal{B}_l=\mathcal{B}_s\cup\mathcal{B}_t$ where $\mathcal{B}_s=\{\{\textbf{s}_{i, j}\}^{N_s}_{i=1}, y_j\}^{M}_{j=1}$ \\ 
     \hspace{5mm}  and $\mathcal{B}_t=\{\{\textbf{t}_{i, j}\}^{N_t}_{i=1}, y_j\}^{M}_{j=1}$. \\ 
    \hspace{1.5mm} 2. Apply strong augmentation on $\mathcal{B}_s$, $\mathcal{B}_t$ and  $\mathcal{B}_u$, \\
    \hspace{5mm} and get $\mathcal{B}'_s=\Phi(\mathcal{B}_s)$, $\mathcal{B}'_t=\Phi(\mathcal{B}_t)$ and $\mathcal{B}^s_u=\Phi(\mathcal{B}_u)$. \\
    \hspace{1.5mm} 3. Apply weak augmentation on $\mathcal{B}_u$ and get  $\mathcal{B}^w_u=\psi(\mathcal{B}_u)$. \\    
    \hspace{1.5mm} 4. Calculate the cross-entropy loss with $\mathcal{B}'_s$ and \\
    \hspace{5mm} $\mathcal{B}'_t$ using Eq. \eqref{loss_ce}. \\
    \hspace{1.5mm} 5. Calculate the unsupervised alignment loss with $\mathcal{B}'_s$ and \\ 
    \hspace{5mm} $\mathcal{B}^w_u$   using an existing UDA method (e.g., MME \cite{saito2019semi}). \\
    \hspace{1.5mm} 6. Calculate the enhanced categorical alignment loss with \\
    \hspace{5mm} $\mathcal{B}'_s$ and $\mathcal{B}'_t$ by using Eq. \eqref{proto_loss_aug} (for the ECACL-P variant), \\
    \hspace{5mm} 	or Eq. \eqref{triplet_loss_aug} (for the ECACL-T variant). \\
    \hspace{1.5mm} 7. Calculate the consistency alignment loss with \\
    \hspace{5mm} $\mathcal{B}^s_u$ and $\mathcal{B}^w_u$ using Eq. \eqref{self_loss}. \\
    \hspace{1.5mm} 8. Optimize the model $h$ using Eq. \eqref{obj}. \\
    \textbf{end while} 
  \\ \hline 
  \end{tabular}
   \vspace{-15pt}
\end{table}

\begin{table*}[t]
\small
    \renewcommand{\tabcolsep}{3.4pt}  
 \begin{center}
 \begin{tabular}{|l|c|cc|cc|cc|cc|cc|cc|cc|cc|} \hline
& \multirow{2}{*}{Net}
&\multicolumn{2}{|c|}{R$\rightarrow$C} &\multicolumn{2}{|c|}{R$\rightarrow$P} & \multicolumn{2}{|c|}{P$\rightarrow$C}  & \multicolumn{2}{|c|}{C$\rightarrow$S} & \multicolumn{2}{|c|}{S$\rightarrow$P} & \multicolumn{2}{|c|}{R$\rightarrow$S} & \multicolumn{2}{|c|}{P$\rightarrow$R}     &\multicolumn{2}{|c|}{Mean} \\ \cline{3-18}
& &1\scriptsize{-shot}&3\scriptsize{-shot} &1\scriptsize{-shot}&3\scriptsize{-shot}&1\scriptsize{-shot}&3\scriptsize{-shot} &1\scriptsize{-shot}&3\scriptsize{-shot}&1\scriptsize{-shot}&3\scriptsize{-shot}&1\scriptsize{-shot}&3\scriptsize{-shot} &1\scriptsize{-shot}&3\scriptsize{-shot} &1\scriptsize{-shot}&3\scriptsize{-shot}  \\ \hline
ST  & AlexNet  & 43.3      & 47.1 & 42.4   & 45.0 & 40.1   & 44.9 & 33.6   & 36.4 & 35.7   & 38.4 & 29.1 & 33.3 & 55.8   & 58.7 & 40.0 & 43.4 \\
 DANN   & AlexNet   & 43.3      & 46.1 & 41.6   & 43.8 & 39.1   & 41.0 & 35.9   & 36.5 &36.9   & 38.9 & 32.5 & 33.4 & 53.6   & 57.3 & 40.4 & 42.4 \\
 ADR   & AlexNet    & 43.1      & 46.2 &    41.4    & 44.4 &    39.3    & 43.6 & 32.8        &   36.4   &  33.1      &  38.9    &   29.1   &  32.4    & 55.9  & 57.3 & 39.2  & 42.7 \\
 CDAN   & AlexNet       & 46.3      & 46.8 & 45.7   & 45.0 & 38.3   & 42.3 & 27.5   & 29.5 & 30.2   & 33.7 & 28.8 & 31.3 & 56.7   & 58.7 & 39.1 & 41.0 \\
 ENT    & AlexNet       & 37.0      & 45.5 & 35.6   & 42.6 & 26.8   & 40.4 & 18.9   & 31.1 & 15.1   & 29.6 & 18.0 & 29.6 & 52.2   & 60.0 & 29.1 & 39.8 \\
 MME    & AlexNet       & 48.9      & 55.6 & 48.0   & 49.0 & 46.7   & 51.7 & 36.3   & 39.4 & 39.4   & 43.0 & 33.3 & 37.9 & 56.8   & 60.7 & 44.2 & 48.2 \\
 Meta-MME & AlexNet & -  & 56.4  & -  & 50.2  & -  & 51.9  & -   & 39.6  & -  & 43.7 & - & 38.7  & -  & 60.7  & -  & 48.7 \\
BiAT & AlexNet  & 54.2 & 58.6 & 49.2 & 50.6 & 44.0 & 52.0 & 37.7 & 41.9 & 39.6 & 42.1 & 37.2 & 42.0 & 56.9 & 58.8 & 45.5 & 49.4 \\
FAN & AlexNet           & 47.7  & 54.6 & 49.0 & 50.5 & 46.9 & 52.1 & 38.5 & 42.6 & 38.5 & 42.2 & 33.8 & 38.7 & 57.5 & 61.4 & 44.6 & 48.9 \\\hline
ECACL-T     & AlexNet      & \bf{56.8} & \bf{62.9}    &\bf{54.8}  & 58.9   & \bf{56.3}  &\bf{60.5}   & \bf{46.6} & \bf{51.0} & \bf{54.6} & \bf{51.2}     & \bf{45.4} & \bf{48.9}   & \bf{62.8} & \bf{67.4}    & \bf{53.4} & \bf{57.7} \\
ECACL-P     & AlexNet       & 55.8  & 62.6  & 54.0  & \bf{59.0}   & 56.1   &  \bf{60.5}   & 46.1  & 50.6 & \bf{54.6} & 50.3  & 45.0  & 48.4   & 62.3 & \bf{67.4}  & 52.8  & 57.6 \\ \hline\hline
ST  & ResNet-34  & 55.6 & 60.0   & 60.6 & 62.2   & 56.8 & 59.4   & 50.8 & 55.0   & 56.0 & 59.5 & 46.3 & 50.1   & 71.8 & 73.9 & 56.9 & 60.0 \\
DANN  & ResNet-34  & 58.2 & 59.8   & 61.4 & 62.8   & 56.3 & 59.6   & 52.8 & 55.4   & 57.4 & 59.9 & 52.2 & 54.9   & 70.3 & 72.2 & 58.4 & 60.7 \\
ADR   & ResNet-34 & 57.1 & 60.7 & 61.3 & 61.9 & 57.0 & 60.7 & 51.0 & 54.4 & 56.0 & 59.9 & 49.0 & 51.1 & 72.0 & 74.2 & 57.6 & 60.4  \\
CDAN  & ResNet-34  & 65.0 & 69.0   & 64.9 & 67.3   & 63.7 & 68.4   & 53.1 & 57.8   & 63.4 & 65.3 & 54.5 & 59.0   & 73.2 & 78.5 & 62.5 & 66.5 \\
ENT   & ResNet-34 & 65.2 & 71.0   & 65.9 & 69.2   & 65.4 & 71.1   & 54.6 & 60.0   & 59.7 & 62.1 & 52.1 & 61.1   & 75.0 & 78.6 & 62.6 & 67.6 \\
MME   & ResNet-34 & 70.0 & 72.2   & 67.7 & 69.7   & 69.0 & 71.7   & 56.3 & 61.8   & 64.8 & 66.8 & 61.0 & 61.9   & 76.1 & 78.5 & 66.4 & 68.9\\
MME   & ResNet-34 & - & 73.5  & -  & 70.3 & - & 72.8  & -  & 62.8  & - & 68.0  & -  & 63.8 & - & 79.2  & -  & 70.1 \\
BiAT & ResNet-34 & 73.0 & 74.9 & 68.0 & 68.8 & 71.6 & 74.6 & 57.9 & 61.5 & 63.9 & 67.5 & 58.5 & 62.1 & 77.0 & 78.6 & 67.1 & 69.7 \\
FAN   & ResNet-34  &70.4 & 76.6 & 70.8 & 72.1 & 72.9 & 76.7 & 56.7 & 63.1 & 64.5 & 66.1 & 63.0 & 67.8 & 76.6 & 79.4 & 67.6 & 71.7 \\\hline
ECACL-T  & ResNet-34 & 73.5    & 76.4 & 72.8 &74.3   &72.8 & 75.9   & \bf{65.1} &65.3    &70.3 &72.2    &64.8  &68.6  &78.3 &79.7  &71.1 &73.2 \\
ECACL-P  & ResNet-34 & \bf{75.3} & \bf{79.0} & \bf{74.1} & \bf{77.3} & \bf{75.3} & \bf{79.4} & 65.0 & \bf{70.6} & \bf{72.1} & \bf{74.6}  & \bf{68.1} & \bf{71.6} & \bf{79.7} & \bf{82.4}  & \bf{72.8} & \bf{76.4}\\ \hline
 \end{tabular}
 \end{center}
 \vspace{-0.1cm}
 \caption{Results on the \textit{DomainNet} dataset. Best results are in \textbf{bold}.
 }
 \label{result_domain_net}
  \vspace{-0.2cm}
\end{table*}

\section{Experiments}
\noindent\textbf{Datasets and evaluation protocols}.
We conduct experiments on three commonly used datasets, namely, 
\textit{VisDA2017}~\cite{peng2017visda}, \textit{DomainNet}~\cite{peng2019moment}, and \textit{Office-Home}~\cite{venkateswara2017Deep}.

\textit{DomainNet} consists of 6 domains of 345 categories. Following \cite{saito2019semi}, we select the \textit{Real} (R), \textit{Clipart} (C), \textit{Painting} (P), and \textit{Sketch} (S) as the 4 evaluation domains and perform the following cross-domain evaluations: R$\rightarrow$C (adaptation from source $\textit{Real}$ to target \textit{Clipart}), R$\rightarrow$P, P$\rightarrow$C, C$\rightarrow$S, S$\rightarrow$P, R$\rightarrow$S, and P$\rightarrow$R. For each set of cross-domain experiment, we evaluate both 1-shot and 3-shot settings, where there are 1 and 3 labeled target samples, respectively. The labeled samples are randomly selected and we use the provided splits for experiments. We evaluate the classification accuracy for all the 7 sets of experiments and also report the mean of the accuracies. 

\textit{VisDA2017} includes 152,397 synthetic and 55,388 real images from 12 categories. 
For SSDA evaluation, we randomly select 1 and 3 real images from each of the 12 classes as the landmarks, which correspond to the 1-shot and 3-shot evaluation settings, respectively. Following precious method  \cite{xu2019larger}, we report the per-class classification accuracy and also the Mean Class Accuracy (MCA) over all classes. 

\textit{Office-Home} contains images of 65 categories that are from 4 different domains, namely, \textit{Real} (R), \textit{Clipart} (C), \textit{Art} (A), and \textit{Product} (P).  We use the same 1-shot and 3-shot splits as \cite{saito2019semi} and evaluate the adaptation performance for all 12 pairs of domains. We report the classification accuracies for all the experimental sets and the accuracy mean. 


\noindent
\textbf{Implementation details.} 
The proposed ECACL-P and ECACL-T are general SSDA frameworks that can incorporate most existing UDA methods and leverage landmark samples to improve adaptation performance.  Depending on the UDA method built upon, we can get different variants.  But for ease and fairness of evaluation, we conduct most of our experiments with the variants based on 
MME \cite{saito2019semi}\footnote{Unless otherwise specified, we use ``ECACL-P'' and ``ECACL-T'' to represent the variants in short.}. Note that although MME is proposed for SSDA, it can be viewed as an UDA method that naively merges labeled target samples into labeled source samples for the cross-entropy loss optimization. Since we do this in the same way, MME is still compatible with our framework. 
We adopt exactly the same training procedures and hyper-parameters as MME, except the way to sample labeled data in each mini-batch. Rather than naively sampling data across the whole labeled set,  we perform class-balanced sampling: In each mini-batch, we randomly sample $M$ classes with $N_s$ and $N_t$ images for each class from source and target domains, respectively. We set $M=$10, $N_s=$10, $N_t=$1 for 1-shot setting, and $N_t=$3 for 3-shot setting.
We set the balancing hyper-parameters of different losses as $\lambda_1=0.1$ and $\lambda_2=1$ in  Eq. (\ref{obj}), and the confident threshold $\sigma=0.8$ for pseudo labeling (Eq. (\ref{self_loss})) for all experiments. For ECACL-T, we set the margin parameter $m=1.0$ for all experiments.

\begin{table*}[t] \renewcommand{\tabcolsep}{6pt}  
\small
    \begin{center}
            \begin{tabular}{|c|cccccccccccc|c|}\hline      
                Method  & plane & bcycl & bus & car & horse & knife & mcycl & person & plant & sktbrd & train & truck & MCA \\\hline            
        \multicolumn{14}{|c|}{\bf{1-shot}} \\\hline                             
                ST     & 82.6 & 52.8 & 75.0 & 57.6 & 72.7 & 39.7 & 80.5 & 53.3 & 59.0 & 64.1 & 77.5 & 12.6 & 60.6 \\    
                MME   & 86.6 & 60.1 & 80.8 & 61.9 & 84.0 & 69.6 & 87.0 & 72.4 & 73.0 & 50.9 & 79.4 & \textbf{14.7} & 68.4  \\\hline   
                ECACL-P   & \textbf{94.9}  & \textbf{81.5}  & \textbf{88.9}  & \textbf{81.3}  & \textbf{95.9}  & \textbf{92.4}  & \textbf{92.2}  & \textbf{83.3}  & \textbf{95.2} & \textbf{77.4} & \textbf{88.4} & 2.3 & \textbf{81.1} \\ \hline
        \multicolumn{14}{|c|}{\bf{3-shot}} \\\hline       
        ST    & 74.0 & 71.7 & 71.2 & 64.7 & 78.5 & 71.8 & 69.6 & 51.4 & 73.7 & 49.4 & 80.8 & 19.8 & 64.7 \\ 
        MME   &87.2 &67.3 &74.9 &64.5 &86.9 &85.5 &78.8 &75.8 &84.4 &48.0 &80.8 &\textbf{19.9} &71.2 \\ \hline
        ECACL-P   & \textbf{95.9}  & \textbf{82.9}  & \textbf{88.6}  & \textbf{84.9}  & \textbf{95.9}  & \textbf{92.1} & \textbf{93.3}  & \textbf{83.7}  & \textbf{95.4}  & \textbf{79.3}  & \textbf{88.0}  & 19.5  & \textbf{83.3} \\\hline               
            \end{tabular}
    \end{center}
     \vspace{-0.1cm}
    \caption{Results on the \textit{VisDA2017} dataset. 
    }
    \label{tab:acc_visda}
     \vspace{-0.2cm}
\end{table*}

\begin{table*}[t]\renewcommand{\tabcolsep}{4pt}  
\small
\begin{center}
\begin{tabular}{|l|cccccccccccc|c|}
\hline
&R to C& R to P & R to A & P to R & P to C & P to A & A to P & A to C & A to R & C to R & C to A & C to P & Mean \\\hline
   \multicolumn{14}{|c|}{\bf{One-shot}} \\\hline
S+T  & 37.5 & 63.1 & 44.8 & 54.3 & 31.7 & 31.5 & 48.8 & 31.1 & 53.3 & 48.5 & 33.9 & 50.8 & 44.1 \\
DANN & 42.5 & 64.2 & 45.1 & 56.4 & 36.6 & 32.7 & 43.5 & 34.4 & 51.9 & 51.0 & 33.8 & 49.4 & 45.1 \\
ADR  & 37.8 & 63.5 & 45.4 & 53.5 & 32.5 & 32.2 & 49.5 & 31.8 & 53.4 & 49.7 & 34.2 & 50.4 & 44.5 \\
CDAN & 36.1 & 62.3 & 42.2 & 52.7 & 28.0 & 27.8 & 48.7 & 28.0 & 51.3 & 41.0 & 26.8 & 49.9 & 41.2 \\
ENT  & 26.8 & 65.8 & 45.8 & 56.3 & 23.5 & 21.9 & 47.4 & 22.1 & 53.4 & 30.8 & 18.1 & 53.6 & 38.8 \\
MME  & 42.0 & 69.6 & 48.3 & 58.7 & 37.8 & 34.9 & 52.5 & 36.4 & 57.0 & 54.1 & 39.5 & 59.1 & 49.2 \\\hline
ECACL-P  & \bf{50.3}   & \bf{70.71}    & \bf{52.2} & \bf{61.4} & \bf{41.2} & \bf{39.3} & \bf{57.8} & \bf{39.1} & \bf{59.1} & \bf{55.8} & \bf{41.7} & \bf{59.9} & \bf{52.4}  \\\hline
\multicolumn{14}{|c|}{\bf{Three-shot}} \\\hline
S+T          & 44.6   & 66.7   & 47.7   & 57.8   & 44.4   & 36.1   & 57.6   & 38.8   & 57.0   & 54.3   & 37.5   & 57.9   & 50.0 \\       
DANN         & 47.2   & 66.7   & 46.6   & 58.1   & 44.4   & 36.1   & 57.2   & 39.8   & 56.6   & 54.3   & 38.6   & 57.9   & 50.3 \\
ADR          & 45.0   & 66.2   & 46.9   & 57.3   & 38.9   & 36.3   & 57.5   & 40.0   & 57.8   & 53.4   & 37.3   & 57.7   & 49.5 \\
CDAN         & 41.8   & 69.9   & 43.2   & 53.6   & 35.8   & 32.0   & 56.3   & 34.5   & 53.5   & 49.3   & 27.9   & 56.2   & 46.2 \\
ENT          & 44.9   & 70.4   & 47.1   & 60.3   & 41.2   & 34.6   & 60.7   & 37.8   & 60.5   & 58.0   & 31.8   & 63.4   & 50.9 \\
MME & 51.2   & 73.0   & 50.3   & 61.6   & 47.2   & 40.7   & 63.9   & 43.8  & 61.4   & 59.9   & 44.7   & 64.7   & 55.2 \\
FAN & 51.9 & 74.6 & 51.2 & 61.6 & 47.9 & 42.1 & 65.5 & 44.5 & 60.9 & 58.1 & 44.3 & 64.8 & 55.6 \\\hline
ECACL-P  & \bf{55.4}   & \bf{75.7} & \bf{56.0} & \bf{67.0} & \bf{52.5} & \bf{46.4} & \bf{67.4} & \bf{48.5} & \bf{66.3} & \bf{60.8} & \bf{45.9} & \bf{67.3} & \bf{59.1}  \\\hline 
\end{tabular}
\end{center}
\vspace{-0.2cm}
\caption{Results on the \textit{Office-Home} dataset.}
\label{tb:office-home_all}
\vspace{-0.1cm}
\end{table*}

\subsection{Comparative Results}
We compare with the following methods,  DANN \cite{DBLP:journals/jmlr/GaninUAGLLML16}, ADR \cite{saito2017adversarial}, CDAN \cite{long2018conditional}, ENT \cite{grandvalet2005semi},  MME \cite{saito2019semi},  FAN \cite{kim2020attract}, BiAT \cite{jiangbidirectional}, and Meta-MME \cite{li2020online}. All these methods are either specifically designed or tailored to address the SSDA problem. We also report results of the baseline method ``ST'' which trains models with labeled samples from source and target domains, without domain alignment. 

\noindent\textit{\textbf{DomainNet}}.
We first compare the two variants of ECACL. We can see from Table \ref{result_domain_net} that while ECACL-T performs slightly better than ECACL-P with the AlexNet backbone, ECACL-P gets much better performance than ECACL-T with the ResNet-34 backbone. 
This shows that the prototype based categorical domain alignment technique is more effective than the triplet loss based one. One possible reason is that the former considers discrimination over all classes, which is more beneficial than the latter where only the triplet-wise relationship is modeled. For ease of evaluation, we only evaluate ECACL-P for the rest experiments. 

Comparing with other methods, we can see that ECACL-P shows significant advantages for all the experimental settings. Specifically, with AlexNet  as the feature extraction model, ECACL-P attains 8.6 and 9.4 point gains for the 1-shot and 3-shot settings, respectively, over MME which ECACL-P is based on. With ResNet-34, the improvements are 6.4 and 7.5 for the 1-shot and 3-shot settings, respectively. ECACL-P also performs significantly better than the most recent methods in both settings with both backbone networks. 
These results substantiate the effectiveness of ECACL-P on mitigating domain shifts by comprehensively exploring  landmarks. 

\noindent\textit{\textbf{VisDA2017}}. We can see from Table \ref{tab:acc_visda} that ECACL-P, built upon on MME, reaches significant gains over MME. The average gains are 12.7 for the 1-shot setting and 12.1 for the 3-shot setting. These tremendous improvements convincingly evidence the effectiveness of ECACL-P.

\noindent\textit{\textbf{Office-Home}}.  We can see from Table \ref{tb:office-home_all} that ECACL-P attains remarkable performance boosts over existing methods as well, although the gains are not as significant as those on the other two datasets. A possible reason is that this dataset is harder than the other two such that improvements are more difficult to attain.

\begin{table}[t]
\vspace{-8pt}
\small
\begin{center}
\renewcommand{\tabcolsep}{3pt}
\begin{tabular}{|c|c|c|c|c|c|c|c|c|}\hline
CA          & &  $\checkmark$    &           &           & $\checkmark$   &   $\checkmark$  &            &  $\checkmark$         \\
SA          & &               & $\checkmark$  &             & $\checkmark$   &              &   $\checkmark$   &  $\checkmark$         \\
CONA        & &               &           & $\checkmark$  &              & $\checkmark$  &   $\checkmark$   &  $\checkmark$         \\\hline
   & 37.9 &    39.2        &  39.3        &  43.5           &     42.2      &  45.2         &   44.1           & 48.4            \\\hline
\end{tabular}
\end{center}
\vspace{-0.2cm}
    \caption{Ablation study for the adaptation from \textit{Real} to \textit{Sketch} on the \textit{DomainNet} dataset for the 3-shot setting.
    The second column shows the baseline result obtained by MME.
    }
    \label{table_ablation_final}
    \vspace{-0.3cm}
\end{table}

\begin{figure*}
\centering
\includegraphics[width=0.85\linewidth]{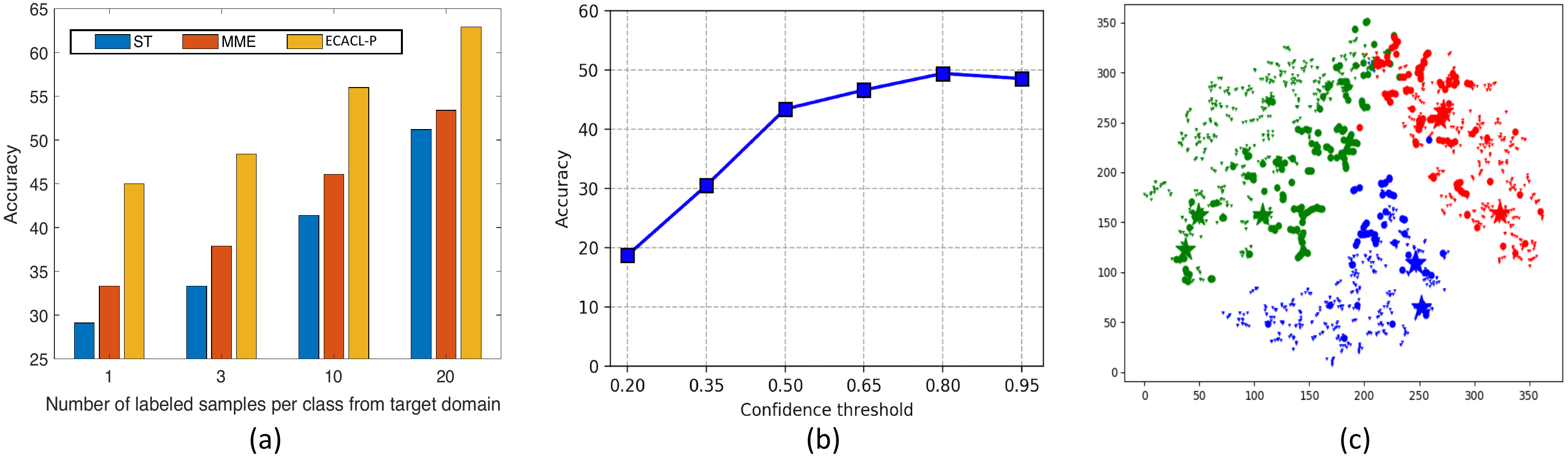}
\vspace{-0.15cm}
\caption{(a): Accuracy with different numbers of labeled samples per class in target domain. (b): Sensitivity analysis with respect to the confident threshold $\sigma$ (Eq. (\ref{self_loss})).
(c): t-SNE visualization of the learned features of 3 randomly selected classes. Different colors represent different classes. Shapes ``Y'', ``{\Large$\star$}'' and 
``$\circ$''
represent source samples, labeled target samples, and unlabeled target samples, respectively. }
\label{further_analysis}  
\vspace{-0.1cm}
\end{figure*}


\subsection{Additional Empirical Analysis}
\noindent\textbf{Ablation study}.
Built upon existing UDA methods, ECACL-P includes the following new modules/techniques to address the SSDA problem, namely,
the Prototypical Alignment (PA) module, the Strong Augmentation (SA) technique that aims to enhance PA and the CONsistency Alignment (CONA) module. Table \ref{table_ablation_final} shows the ablation study on the adaptation from \textit{Real} to \textit{Sketch} on the \textit{DomainNet} dataset for the 3-shot setting. 
We can see that all the new modules/techniques contribute to the ultimate performance promotions, thus verifying the efficacy.

\begin{table}
\small
 \vspace{-0.3cm}
   \renewcommand{\tabcolsep}{3pt}
   \begin{center}
         \begin{tabular}{|c|c|c|cc|}\hline      
            Method            & Net & Setting & \multicolumn{2}{|c|}{MCA} \\\hline     
            Source-only    & ResNet-101 & UDA & \multicolumn{2}{|c|}{52.4}  \\
            HAFN   & ResNet-101 & UDA& \multicolumn{2}{|c|}{73.9}  \\
            SAFN  & ResNet-101 & UDA & \multicolumn{2}{|c|}{76.1}  \\ \hline
         & & & 1-shot & 3-shot \\\hline
        HAFN + ST  & ResNet-101 & SSDA   & 77.0 &  79.3 \\  
            SAFN + ST  & ResNet-101  & SSDA  & 77.5 &  79.2\\ 
            ECACL-P (HAFN)  & ResNet-101  & SSDA & \textbf{83.9} & \textbf{85.3} \\             
            ECACL-P (SAFN)  & ResNet-101  & SSDA & 83.3  & 84.5 \\\hline               
        \hline                
            ST & ResNet-34   & SSDA & 60.6  & 64.7 \\ 
            MME  & ResNet-34  & SSDA & 68.4   &71.2 \\   
            ECACL-P (MME)  & ResNet-34 & SSDA  & \textbf{81.1} & \textbf{83.3} \\ \hline        
         \end{tabular}
   \end{center}
    \vspace{-0.14cm}
   \caption{Flexibility analysis of ECACL-P on the \textit{VisDA2017} dataset. 
   ``HAFN + ST'' and ``SAFN + ST'' denote the naive extensions of the methods HAFN and SAFN from UDA to SSDA, by further including labeled target data for the cross-entropy loss optimization.
   ``ECACL-P (MME)'', ``ECACL-P (HAFN)'', and ``ECACL-P (SAFN)'' are the methods corresponding to different UDA methods incorporated into our framework.
   }
   \label{tab:acc_visda_all}
    \vspace{-0.18cm}
\end{table}

\noindent\textbf{Plug-and-play evaluation}. As mentioned above, ECACL-P is agnostic to the UDA methods built upon. To evaluate this, we apply ECACL-P as a plug-and-play component on different existing UDA methods and see how much adaption performance can be improved.  Table \ref{tab:acc_visda_all} shows the results of three UDA methods before and after empowered by ECACL-P, namely HAFN \cite{xu2019larger}, SAFN \cite{xu2019larger}, and MME \cite{saito2019semi}. 
We can see from Table \ref{tab:acc_visda_all} that ECACL-P indeed significantly promotes the performance of existing UDA methods and their naive SSDA extensions. 
For example, ``ECACL-P (HAFN)'' raises the result of the UDA method HAFN from 73.9 to 83.9 with one sample per class labeled and further to 85.3 with 3 samples per class labeled. 
These significant and consistent improvements with different UDA methods, different backbone networks and different numbers of landmarks convincingly substantiate the effectiveness of ECACL-P on boosting adaptation performance with minimal labeling effort. 

\noindent\textbf{Impact of the number of landmarks}.
We have shown the superior performance of the standard 1-shot and 3-shot settings above. 
We wonder how performance changes with the number of landmarks increased. 
We study this on the adaptation from \textit{Real} to \textit{Sketch} on \textit{DomainNet}. 
We can see from Fig. \ref{further_analysis} (a) that all the three methods enjoy performance boosts with more target samples labeled. Comparatively speaking, ECACL-P consistently reaches the best performance for all the cases. This substantiates the benefit of our method for flexibly exploiting different amount of labeled target samples to help address the domain shift problem. 


\noindent\textbf{Parameter analysis}.
Our consistency alignment module involves an important hyper-parameter 
$\sigma$ which is the confident threshold for pseudo labeling.  
We analyze the sensitiveness regarding $\sigma$ on the adaptation from  \textit{Real} to \textit{Sketch} on the \textit{DomainNet} dataset under the 3-shot setting. The result is shown in Figure \ref{further_analysis} (b). We see that ECACL-P is not very sensitive to $\sigma$ and maintains good performance when $\sigma$ is fairly high (over 0.65).

\begin{table}[t]
		\vspace{-8pt}
	\small
    \renewcommand{\tabcolsep}{5pt}
\begin{center}
\begin{tabular}{|l|c|c|c|c|}\hline
						   &	   & Split 1  & Split 2 	& Split 3 \\\cline{1-5}    
\multirow{2}{*}{AlexNet}   & MME   & 37.9  	  & 41.2		& 43.0						\\
                           & ECACL-P  & 48.4  	  & 48.9		& 51.6					\\\cline{1-5}
\multirow{2}{*}{ResNet-34} & MME   & 61.9  	  & 65.3		& 65.2					\\ 
                           & ECACL-P  & 71.6  	  & 72.5		& 71.8					\\\hline
\end{tabular}           
\end{center}
\vspace{-0.26cm}
    \caption{Variance analysis with different dataset splits.}
    \label{table_splits}
    \vspace{-0.5cm}
\end{table}

\noindent\textbf{Results with different splits}.
We follow the prior method \cite{saito2019semi} and use the provided labeled/unlabeled splits of target domains for experiments. 
To study the impact of randomness on the performance, we regenerate the splits used for the adaptation from the \textit{real} domain to \textit{sketch} domain in the $\textit{DomainNet}$ dataset for the 3-shot settings. Table \ref{table_splits} shows the results with  three different splits. We can see that the proposed ECACL-P consistently improves MME with different splits.

\noindent\textbf{Feature visualization}.
To qualitatively evaluate the alignment results, we plot the t-SNE \cite{maaten2008visualizing} visualization of the features produced by ECACL-P for the adaptation from \textit{Real} to \textit{Sketch} on \textit{DomainNet} in the 3-shot setting.  We can see from Figure \ref{further_analysis}(c) that the learned features exhibit favorable clustering structure. Features from different domains are close if they belong to the same classes and apart otherwise. 
This plot further supports that both feature discriminabilty and domain alignment are achieved by the learned model. 

\section{Conclusion}
We propose in this paper a novel semi-supervised domain adaptation (SSDA) framework within which existing unsupervised domain adaptation (UDA) methods can effectively utilize a few labeled samples from the target domain to further mitigate domain shifts. The proposed framework includes two categorical alignment techniques, both of which are further enhanced by a data augmentation based technique that produces highly perturbed images to mitigate overfitting. A consistency alignment module is incorporated into the framework which enforces consistency regularization on the learned model. Ablation study verifies the contributing roles of all the above alignment components. Experiments show that the proposed framework reaches state-of-the-art SSDA performance and consistently promotes the adaptation performance of various UDA methods for different numbers of labeled target samples.

\vspace{5pt}
\noindent\textbf{Acknowledgement}: 
The work was partially supported by the National Science Foundation Award ECCS-1916839.

\section{Appendix}

\begin{figure*}
\centering
\includegraphics[width=1.0\linewidth]{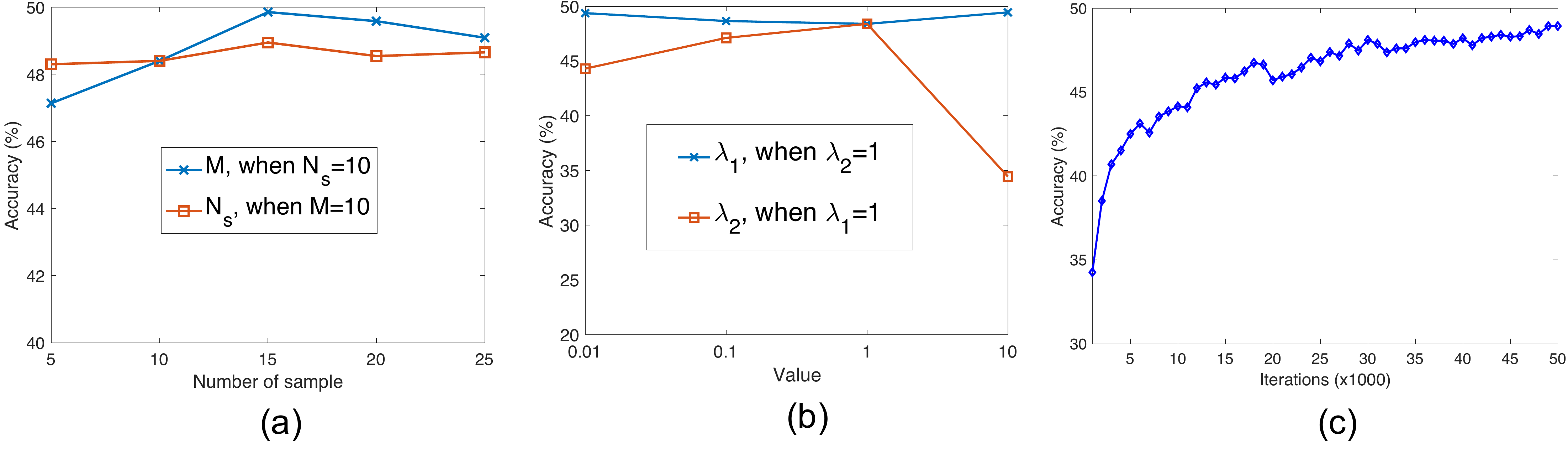}
\vspace{-0.8cm}
\caption{Further analysis of the proposed method. (a): Accuracy with respect to number of classes and number of samples per class in each mini-batch; (b): Accuracy with respect to the two hyper-parameters for different losses; (c): Accuracy with respect to training iterations.
All experiments are conducted on the adaptation from $\textit{Real}$ to $\textit{Sketch}$ in the $\textit{DoaminNet}$ dataset.
}
\label{further_analysis_1}  
\end{figure*}

\subsection{Analysis of the Sampling Strategy}
We perform class-balanced sampling for labeled data: In each mini-batch, we sample $M$ classes with $N_s$ source images and  $N_t$ target images for each class, which results in a mini-batch of size $M*(N_s+N_t)$. We study the impact of batch size to the performance by varying these numbers. Since $N_t$ is a constant for the 1-shot setting ($N_t=1$) and 3-shot setting ($N_t=3$), we only vary the values of $M$ and $N_s$. When evaluating one parameter, we fix the other one. The results of applying ECACL-P on top of MME \cite{saito2019semi} for the adaptation from $\textit{Real}$ to $\textit{Sketch}$ in the $\textit{DoaminNet}$ dataset are shown in Figure \ref{further_analysis_1}(a). We can see from this figure that the proposed method is quite robust with the two hyper-parameters. There are fairly small changes (less than 2\%) when their values are changed from 5 to 25.


\subsection{Parameter Analysis for Different Losses}
As shown in Eq. \eqref{obj}, the overall learning objective of the proposed framework is a weighted sum of different losses. We conduct experiments to evaluate the sensitiveness of the proposed framework with respect to the hyper-parameters, i.e., $\lambda_1$ and $\lambda_2$. When evaluating one parameter, we vary its value and fix the other parameter unchanged.  Figure \ref{further_analysis_1}(b) shows the results for the adaptation from $\textit{Real}$ and $\textit{Sketch}$ in the $\textit{DoaminNet}$ dataset for the 3-shot setting. We can see that the proposed ECACL-P is quite robust with $\lambda_1$ - the performance is stable when $\lambda_1$ varies in a wide range. ECACL-P is more sensitive to $\lambda_2$ and the performance degrades sharply when $\lambda_2$ is greater than 1.

\subsection{Changes of Accuracy during Training}
Figure \ref{further_analysis_1}(c) shows the curve of accuracy with respect to training steps for the adaptation from the \textit{real} domain  to \textit{sketch} domain in the $\textit{DomainNet}$ dataset. We can see that ECACL-P progressively improves the performance as the training process proceeds.

\begin{table*}[t]
	\begin{center}
		\scalebox{0.85}{
			\begin{tabular}{|c|c|cccccccccccc|c|}\hline      
				Method & Net & plane & bcycl & bus & car & horse & knife & mcycl & person & plant & sktbrd & train & truck & MCA \\\hline		
				\multicolumn{15}{|c|}{\bf{UDA}} \\\hline
				Source-only  & ResNet-101 & 55.1 & 53.3 & 61.9 & 59.1 & 80.6 & 17.9 & 79.7 & 31.2 & 81.0 & 26.5 & 73.5 & 8.5 & 52.4 \\
				DAN  & ResNet-101 & 87.1 & 63.0 & 76.5 & 42.0 & 90.3 & 42.9 & 85.9 & 53.1 & 49.7 & 36.3 & 85.8 & 20.7 & 61.1 \\
				DANN   & ResNet-101 & 81.9 & 77.7 & 82.8 & 44.3 & 81.2 & 29.5 & 65.1 & 28.6 & 51.9 & 54.6 & 82.8 & 7.8 & 57.4 \\
				MCD   & ResNet-101 & 87.0 & 60.9 & 83.7 & 64.0 & 88.9 & 79.6 & 84.7 & 76.9 & 88.6 & 40.3 & 83.0 & 25.8 & 71.9 \\
				HAFN   & ResNet-101 & 92.7 & 55.4 & 82.4 & 70.9  & 93.2 & 71.2 & 90.8 & 78.2 & 89.1 & 50.2 & 88.9 & 24.5 & 73.9 \\
				SAFN  & ResNet-101 &  93.6 & 61.3 & 84.1 & 70.6 & 94.1 & 79.0 & 91.8 & 79.6 & 89.9 & 55.6 & 89.0 & 24.4 & 76.1 \\ \hline
        \multicolumn{15}{|c|}{\bf{SSDA (1-shot)}} \\\hline
        HAFN + ST  & ResNet-101    & 92.6 & 64.6 & 88.0 & 66.3 & 93.6 & 84.6 & 90.8 & \textbf{80.9} & 90.8 & 60.1 & 88.0 & 24.5 & 77.0\\  
				SAFN + ST  & ResNet-101    & 95.5 & 64.0 & 80.1 & 63.2 & 93.0 & 87.3 & 91.1& 78.9 & 90.0 & 66.6 & 89.4 & \textbf{30.6} & 77.5 \\ 
				ECACL-P (HAFN)  & ResNet-101   & \textbf{97.4} & \textbf{78.7} & \textbf{90.0} & 86.6 & \textbf{97.7} & 90.2 & 94.2 & 78.0 & \textbf{93.8} & 80.7 & 93.8 & 25.2 & \textbf{83.9} \\					
				ECACL-P (SAFN)  & ResNet-101   & 96.7 & 76.5 & 88.4 & \textbf{90.3} & 97.1 & \textbf{91.9} & \textbf{95.4} & 74.9 & 91.6 & \textbf{82.9} & \textbf{94.8} & 19.5 & 83.3  \\\hline	
        \multicolumn{15}{|c|}{\bf{SSDA (3-shot)}} \\\hline
        HAFN + ST  & ResNet-101  & 93.8& 77.9& \textbf{87.9} & 68.6& 94.2& 88.8& 92.0& 82.3& 91.1& 62.1& 82.8& 30.6& 79.3 \\  
        SAFN + ST  & ResNet-101   & 95.4 & 75.5 & 83.0 & 67.9 & 94.4 & 87.6 & 88.5 & \textbf{78.6} & 93.2 & 66.7 & 86.1 & 33.9 & 79.2 \\ 
        ECACL-P (HAFN)  & ResNet-101 & \textbf{98.2} & 80.3 & 87.5 & 86.8 & \textbf{97.6} & 86.0 & \textbf{94.0} & 73.8 & \textbf{96.3} & \textbf{91.6} & 94.1 & \textbf{37.6} & \textbf{85.3}  \\          
        ECACL-P (SAFN) & ResNet-101  & 97.6  & \textbf{83.7}  & 86.1 & \textbf{91.7}  & 97.5  & \textbf{91.3}  & 92.5  & 59.1  & 94.9  & 90.9  & \textbf{95.9} & 32.5 & 84.5 \\  
        \hline\hline							
        \multicolumn{15}{|c|}{\bf{SSDA (1-shot)}} \\\hline								
				ST & ResNet-34   & 82.6 & 52.8 & 75.0 & 57.6 & 72.7 & 39.7 & 80.5 & 53.3 & 59.0 & 64.1 & 77.5 & 12.6 & 60.6 \\	
				MME  & ResNet-34  & 86.6 & 60.1 & 80.8 & 61.9 & 84.0 & 69.6 & 87.0 & 72.4 & 73.0 & 50.9 & 79.4 & \textbf{14.7} & 68.4  \\	
				ECACL-P (MME)  & ResNet-34 & \textbf{94.9}  & \textbf{81.5}  & \textbf{88.9}  & \textbf{81.3}  & \textbf{95.9}  & \textbf{92.4}  & \textbf{92.2}  & \textbf{83.3}  & \textbf{95.2} & \textbf{77.4} & \textbf{88.4} & 2.3 & \textbf{81.1} \\ \hline
        \multicolumn{15}{|c|}{\bf{SSDA (3-shot)}} \\\hline       
        ST & ResNet-34   & 74.0 & 71.7 & 71.2 & 64.7 & 78.5 & 71.8 & 69.6 & 51.4 & 73.7 & 49.4 & 80.8 & 19.8 & 64.7 \\ 
        MME  & ResNet-34 &87.2 &67.3 &74.9 &64.5 &86.9 &85.5 &78.8 &75.8 &84.4 &48.0 &80.8 &\textbf{19.9} &71.2 \\ 
        ECACL-P (MME)  & ResNet-34 & \textbf{95.9}  & \textbf{82.9}  & \textbf{88.6}  & \textbf{84.9}  & \textbf{95.9}  & \textbf{92.1} & \textbf{93.3}  & \textbf{83.7}  & \textbf{95.4}  & \textbf{79.3}  & \textbf{88.0}  & 19.5  & \textbf{83.3} \\\hline 				
			\end{tabular}
		}
	\end{center}
	 \vspace{-0.3cm}
	\caption{Complete results for the flexibility analysis. Refer Table \ref{tab:acc_visda_all} for more details.
	}
	\label{tab:acc_visda_all1}
	 \vspace{-0.1cm}
\end{table*}

\subsection{Implementation Details}
Our framework is general such that any existing unsupervised domain adaptation (UDA) method can be incorporated and has the adaptation performance improved. 
We verify this by employing three UDA methods, i.e., MME \cite{saito2019semi}, HAFN \cite{xu2019larger} and SAFN \cite{xu2019larger}, which results in three variants of our methods, ``ECACL-P (MME)", ``ECACL-P (HAFN)", and ``ECACL-P (SAFN)", respectively. 
To make fair comparison, we use the official implementations\footnote{MME:  \url{https://github.com/VisionLearningGroup/SSDA_MME}. HAFN and SAFN: \url{https://github.com/jihanyang/AFN.git}.
} of the three UDA methods and follow strictly the same training rules and procedures, except the mini-batch sampling strategy. We have introduced the implementation details specific to our framework in the main text. Here we give a more complete description about the implementation of each of our variant. 

\noindent\textbf{ECACL-P (MME)}:
Following MME, we use the ImageNet pre-trained models to initialize 
both the AlexNet and Resnet-34 backbones. For AlexNet, the last linear layer are replaced with randomly initialized linear layer. For ResNet34, the last linear layer are replaced with two fully-connected layers. The models are optimized with the momentum optimizer. We set the initial learning rate as 0.01 for all fully-connected layers while use a smaller learning rate 0.001 for other layers including convolution layers and batch-normalization layers. The learning rate decay policy is the same as in \cite{ganin2015unsupervised}. Each mini-batch consists of labeled source, labeled target and unlabeled target images. As mentioned in the main text, we perform class balanced sampling for labeled images. For unlabeled images, following MME, we sample in each mini-batch 24 images for ResNet-34 and 32 for AlexNet. 


\noindent\textbf{ECACL-P (HAFN)}. We initialize our backbone with ImageNet pre-trained model. Following HAFN, we perform two-step training, where in the first step we train the model with only labeled data from the source domain; in the second step, we conduct domain adaptive training. For both steps, SGD is employed for model optimization and the learning rate is $10^{-3}$. We train both stages with 10 epochs. 


\noindent\textbf{ECACL-P (SAFN)}: Everything is same as ``ECACL-P (HAFN)" except the UDA term used. 
So, the implementation details are almost the same as those of ``ECACL-P (HAFN)", except that we perform adaptive training and the model is trained with 50 epochs.

\subsection{Complete Results for the Flexibility Analysis}
In the main text, we showed the plug-and-play evaluation results on the $\textit{VisDA17}$ dataset, which substantiated the flexibility of the proposed framework on improving different existing UDA methods. To save space, we only show the MCA (mean class accuracy) for all the classes in the main text.
Here we show the complete results as shown in Table \ref{tab:acc_visda_all1}.

\balance

{\small
\bibliographystyle{ieee_fullname}
\bibliography{egbib}
}

\end{document}